\documentclass[letterpaper, 10 pt, journal, twoside]{ieeetran}  

\IEEEoverridecommandlockouts                              

\usepackage{amsfonts}
\usepackage{amsmath}
\usepackage{color}
\usepackage{graphicx}
\usepackage{makecell}
\usepackage{lipsum}
\usepackage{subfig}
\usepackage{cite}
\usepackage[hyphens]{url}
\usepackage{hyperref}
\usepackage[font=footnotesize]{caption}
\usepackage{breakurl}
\usepackage{tcolorbox}
\usepackage{tabularx}

\usepackage{acronym}
\newacro{LLM}{large language model}

\newcommand{\rev}[1]{\textcolor{black}{#1}}
\newcommand{\refwithletter}[2][]{%
    \hyperref[#2]{\ref*{#2}#1}%
}

\title{\LARGE \bf SwarmGPT: Combining Large Language Models with Safe Motion Planning for Drone Swarm Choreography}

\author{Martin Schuck, Dinushka Orrin Dahanaggamaarachchi, Ben Sprenger, Vedant Vyas, \\Siqi Zhou, and Angela P. Schoellig

\thanks{Manuscript received: April, 15, 2025; Revised July, 31, 2025; Accepted October, 10, 2025.}
\thanks{This paper was recommended for publication by Editor M. Vincze upon evaluation of the Associate Editor and Reviewers' comments.
This work was supported by the \href{https://www.robotics-institute-germany.de/}{Robotics Institute Germany} under BMBF grant 16ME0997K, by the European Union’s Horizon Europe research and innovation programme under the Marie Skłodowska-Curie grant agreement No. 101155035, and by the Humboldt Professorship for Robotics and Artificial Intelligence.}
\thanks{The authors are with the \href{https://www.learnsyslab.org/}{Learning Systems and Robotics Lab} and the \href{http://mirmi.tum.de/}{Munich Institute of Robotics and Machine Intelligence}, Technical University of Munich, 80333 Munich, Germany
        {\tt\footnotesize firstname.lastname@tum.de}}%
\thanks{Digital Object Identifier (DOI): see top of this page.}
}

\newcommand{\bx}{\mathbf{x}}
\newcommand{\bv}{\mathbf{v}}
\newcommand{\bu}{\mathbf{u}}
\newcommand{\bc}{\mathbf{c}}
\newcommand{\bp}{\mathbf{p}}
\newcommand{\br}{\mathbf{r}}

\newcommand{\bg}{\mathbf{g}}

\newcommand{\Rbb}{\mathbb{R}}
\newcommand{\Nbb}{\mathbb{N}}
\renewcommand{\Bbb}{\mathbb{B}}
\newcommand{\Kbb}{\mathbb{K}}
\newcommand{\Jbb}{\mathbb{J}}

\begin{document}

\maketitle

\begin{abstract}
\rev{Drone swarm performances---synchronized, expressive aerial displays set to music---have emerged as a captivating application of modern robotics. Yet designing smooth, safe choreographies remains a complex task requiring expert knowledge.}
We present SwarmGPT, a language-based choreographer that leverages the reasoning power of large language models (LLMs) to streamline drone performance design. The LLM is augmented by a safety filter that ensures deployability by making minimal corrections when safety or feasibility constraints are violated. \rev{By decoupling high-level choreographic design from low-level motion planning, our system enables non-experts to iteratively refine choreographies using natural language without worrying about collisions or actuator limits.  We validate our approach through simulations with swarms up to 200 drones and real-world experiments with up to 20 drones performing choreographies to diverse types of songs, demonstrating scalable, synchronized, and safe performances. Beyond entertainment, this work offers a blueprint for integrating foundation models into safety-critical swarm robotics applications.}
\end{abstract}


\section{INTRODUCTION}
\IEEEPARstart{A}{utonomous} drones are increasingly featured in large-scale events—from concerts and sports championships to Olympic opening ceremonies~\cite{ackerman_2014,ioc_2021,omahony_2020}. Their agility and ability to move in tightly synchronized formations make them powerful tools for creating striking visual effects and conveying artistic expression alongside human performers.

Despite their potential, designing choreographed movements for drone swarms remains a complex challenge. Performances must not only achieve visual and artistic goals but also satisfy system constraints, such as smoothness, collision avoidance, feasibility, and downwash effects~\cite{schollig2011feasiblity,desai2016dynamically,preiss2017downwash}. Balancing expressiveness with safety and hardware feasibility requires significant expert knowledge, making the design process labour-intensive and inaccessible to non-experts.

\begin{figure}[t]
    \centering
    \subfloat[Long-exposure image of 12 drones transitioning from a helix to a spiral motion at a musical beat.]{\includegraphics[width=\columnwidth, trim={0 5cm 5cm 15cm}, clip]{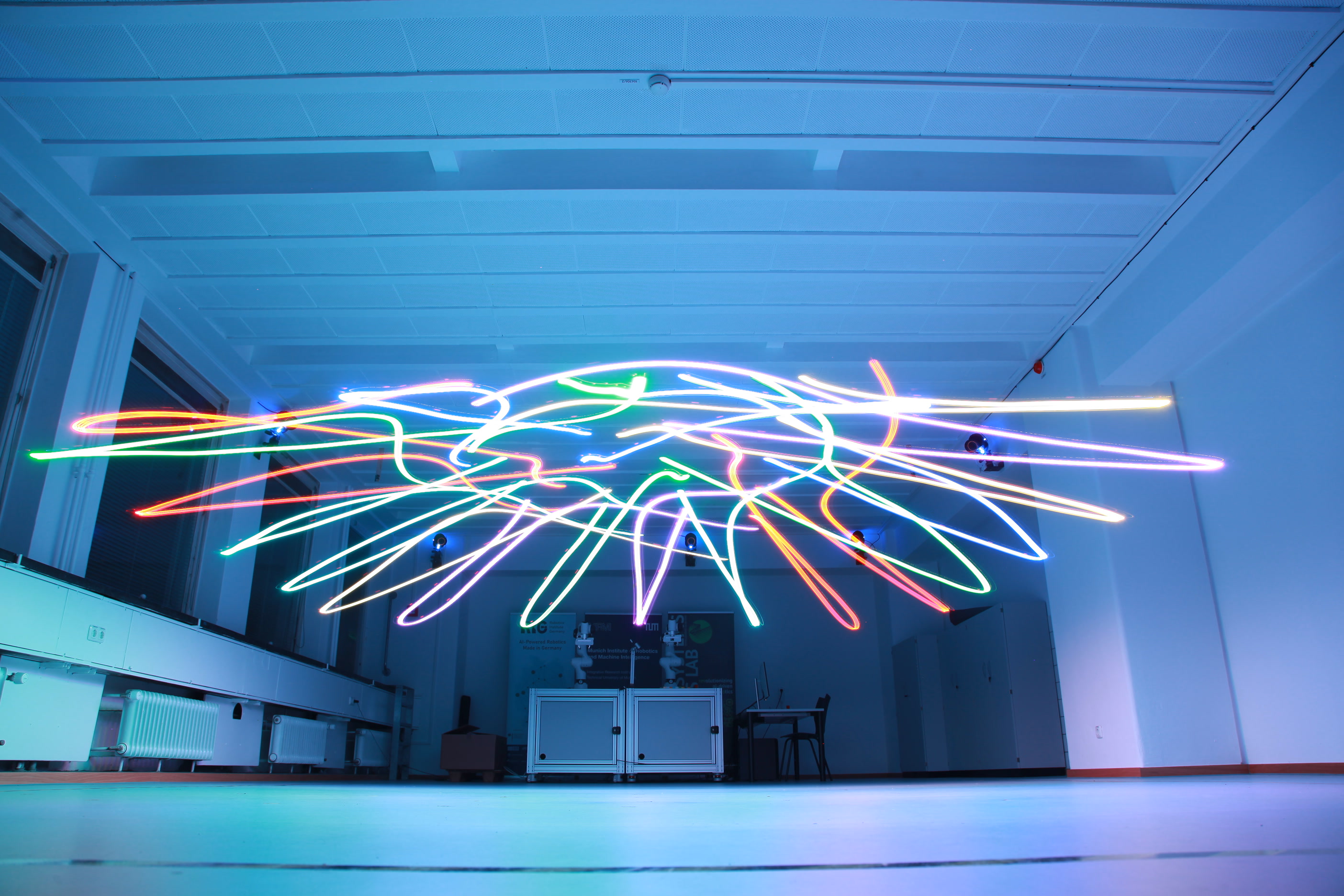}}
    \vfill \vspace{-5pt}
    \subfloat[A close-up view of one of the performing drones.]{\includegraphics[width=\columnwidth, trim={0 5cm 0 15cm}, clip]{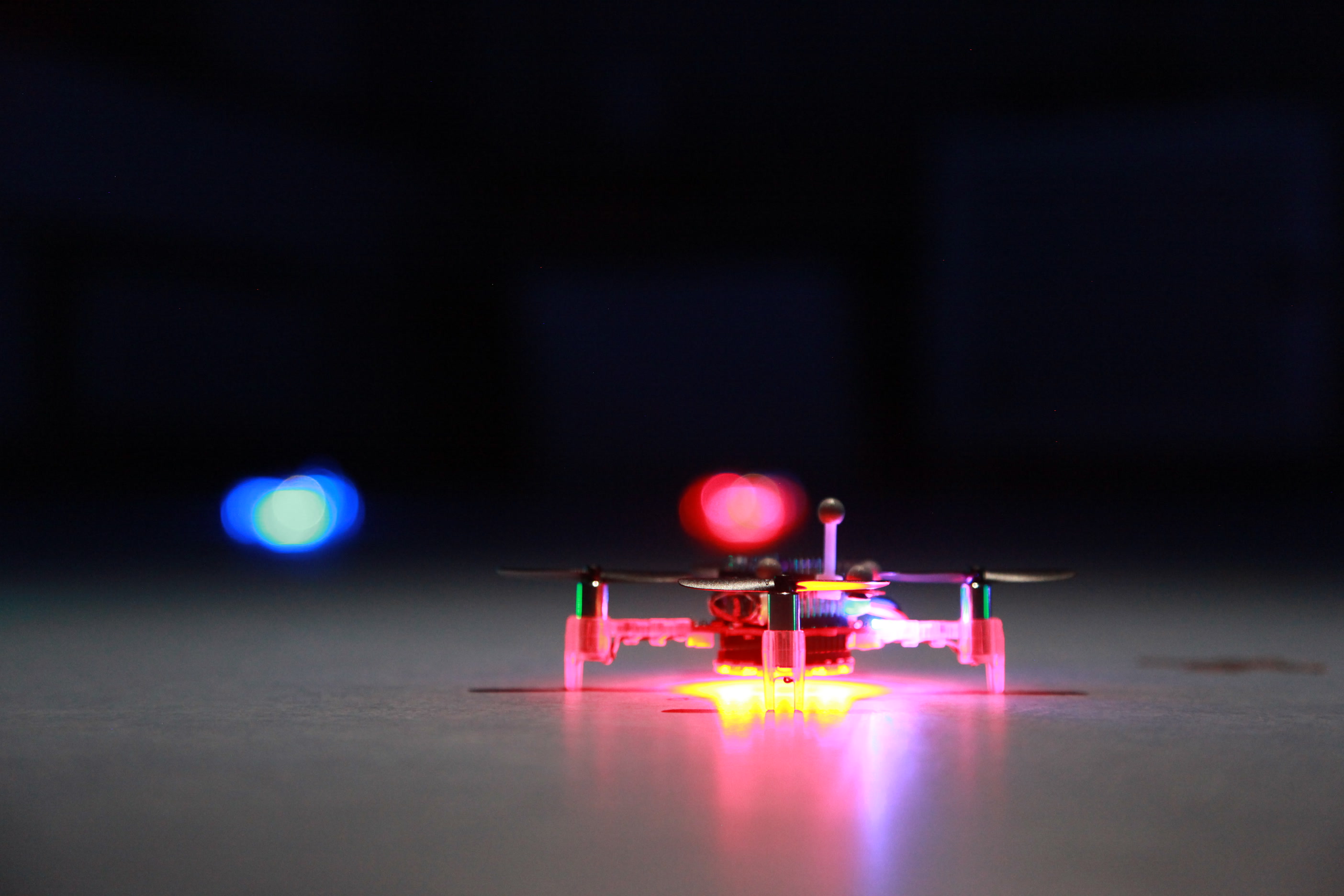}}
    \caption{SwarmGPT generates drone swarm choreographies from natural language, synchronized with music. Demonstration videos are available at \href{https://tiny.cc/swarmgpt}{\url{https://tiny.cc/swarmgpt}} and in the supplementary materials (S1–S7). The project website is \href{https://utiasdsl.github.io/swarm_GPT/}{\url{https://utiasdsl.github.io/swarm_GPT/}}.} 
    \label{fig:intro}
\end{figure}

Recent advances in foundation models, driven by large-scale datasets, have established language and vision-language models as intuitive interfaces for human–robot interaction~\cite{bommasani2021opportunities}. Their semantic understanding and reasoning capabilities have been increasingly applied in robotics—for example, in contextual navigation~\cite{gu2024conceptgraphs,singh2023progprompt}, affordance-driven grasping and manipulation~\cite{rashid2023language,brohan2023can}, and guiding reward design or data collection for skill learning~\cite{ha2023scaling}.
\rev{Language-based control of robot swarms holds similar promise, with potential applications extending beyond performances to domains such as disaster response or search-and-rescue—enabling non-experts to safely and effectively interact with complex systems.}

However, integrating foundation models into real-world robotics remains challenging. Even small errors in perception or reasoning can propagate through control loops, leading to irreversible outcomes~\cite{bommasani2021opportunities,firoozi2023foundation}. These risks are amplified in swarm robotics, and particularly in drone swarms, where any physical contact between agents can cause immediate and catastrophic failure.

\begin{figure*}
    \centering
    \vspace{10pt}
    \includegraphics[width=\textwidth]{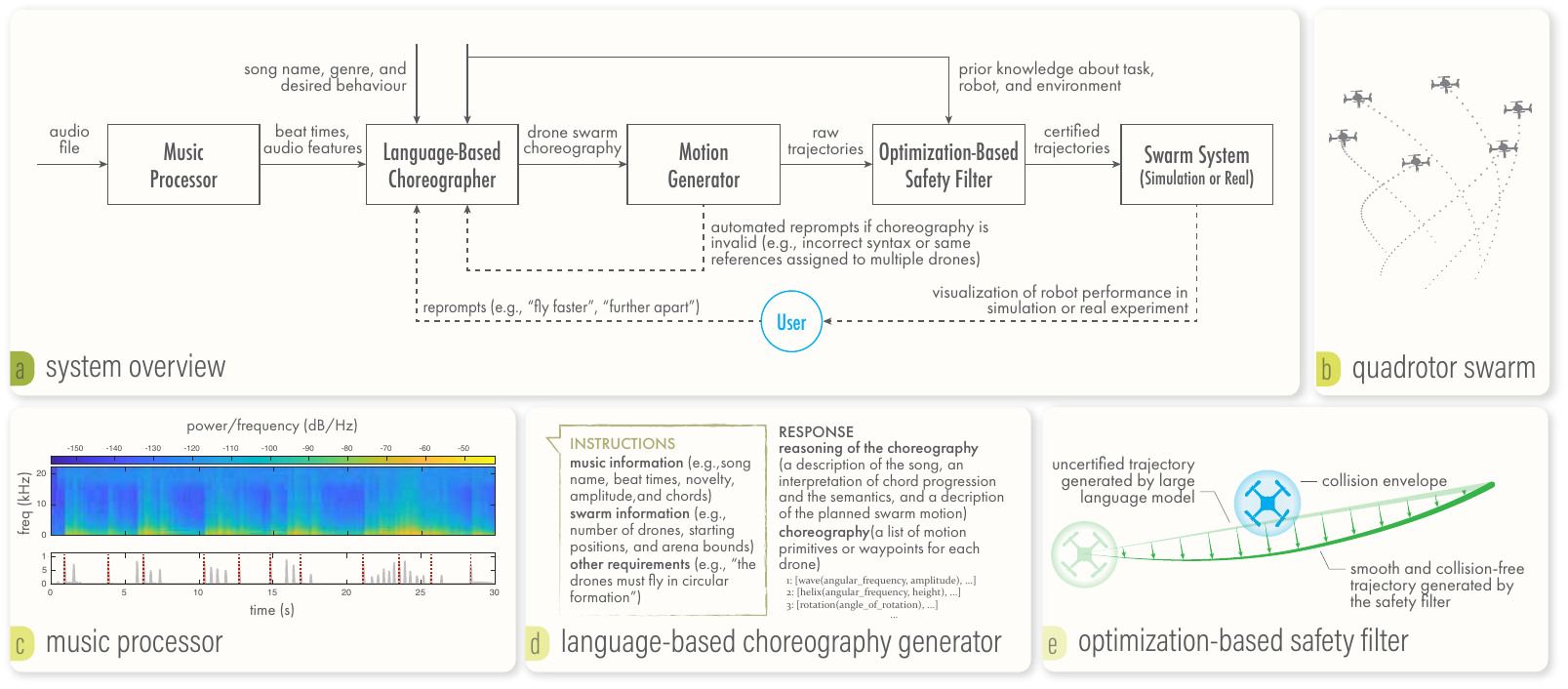}
    \caption{Overview of the SwarmGPT system. \textbf{(a)} A high-level LLM designs unique choreographies, while a low-level optimization-based safety filter ensures feasible, collision-free deployment. \textbf{(b)} The drone performers are quadrotor vehicles. \textbf{(c)}  A music processor extracts beat times and audio features from the song. \textbf{(d)} The LLM generates choreographies from the audio features and music beat times, producing raw drone trajectories. \textbf{(e)} The safety filter outputs safe, feasible trajectories for simulation or hardware deployment, enabling non-experts to iteratively refine via natural language.}
    \label{fig:overview}
\end{figure*}

Building on prior work in drone performances~\cite{Du2018FastAI, schoellig2014so}, we present SwarmGPT—a language-based choreographer that uses the reasoning capabilities of \acp{LLM} to design synchronized, rhythmic drone swarm performances (Fig.~\ref{fig:intro}). To ensure real-world deployability, we introduce a  safety filter that preserves the intended choreography while enforcing collision-free and physically feasible trajectories.
SwarmGPT offers an intuitive interface that enables non-experts to create and iteratively refine drone swarm behaviours using natural language—without needing to account for safety or low-level feasibility constraints.

The key contributions of this work are:
\begin{enumerate}
    \item \textit{SwarmGPT}: A language-based choreographer that combines the reasoning abilities of LLMs with optimization-based safety filters to generate drone swarm performances.
    \rev{\item \textit{Scalability}: Support for large swarms via motion primitives, self-correction mechanisms, and a distributed safety filter optimizer that preserves choreographic intent while ensuring safety and feasibility.}
    \rev{\item \textit{Validation}: Extensive simulations  (up to 200 drones) and real-world performances (up to 20 drones) across multiple musical genres, demonstrating the music interpretation capabilities of the LLM, the safe execution, and the effectiveness of language reprompting.}
\end{enumerate}

\section{Related Work}

\subsection{Robot Choreography}
Robot choreography design to music has emerged as a compelling form of robotic performance, explored across various platforms including humanoids~\cite{boston_dynamics, guizzo2019leaps, xia2012autonomous, boukheddimi2022robot}, quadrupeds~\cite{boston_dynamics, guizzo2019leaps}, robotic arms~\cite{rogel2022robogroove}, and drone swarms~\cite{sparked, schollig2010synchronizing, du2019fast}. Designing expressive and synchronized motions typically involves extensive manual tuning and domain expertise to ensure feasibility, safety, and alignment with musical or visual intent, making the process labour-intensive and non-intuitive.

In aerial swarm choreography, prior work has primarily relied on manually constructed sequences of motion primitives with hand-tuned parameters~\cite{augugliaro_dance_2013,du2019fast}. These approaches require expert involvement to align choreography with musical timing and expression.

In contrast, we propose a language-based choreographer that generates synchronized drone swarm performances from natural language descriptions in minutes. Our system lowers the barrier to creating complex, deployable swarm behaviours by providing an intuitive interface for non-experts.

\subsection{LLMs for Robotics}
Recent work has explored the use of \acp{LLM} for robotic decision-making~\cite{bommasani2021opportunities}, with language guiding movement in tasks such as visual-language navigation and trajectory refinement~\cite{huang2023visual, bucker2023latte}. Beyond navigation, LLMs have been applied to joint navigation and manipulation in unstructured environments~\cite{zitkovich2023rt2}, as well as high-level task planning—whether through code generation~\cite{tang_saytap:_2023}, translating instructions into action sequences~\cite{driess23palme}, or selecting action templates to control robots~\cite{vemprala_chatgpt_2023,singh2023progprompt}.

Building on these directions, we apply LLMs to drone swarm coordination, focusing on language-driven choreography and its safe real-world deployment. A central challenge in robotics is grounding language in the robot’s embodiment and capabilities. Prior work has approached grounding by predicting platform-specific affordances~\cite{brohan2023can}, using self-correcting inner monologues based on feedback~\cite{huang2022inner}, or fine-tuning on physical interactions to improve physical reasoning~\cite{gao2024physically}.

In our approach, grounding is achieved by converting high-level LLM-generated actions into deployable choreographies using a low-level, model-based safety filter that enforces collision avoidance and physical feasibility. 

\subsection{Safe Robot Swarm Decision-Making}
Deploying drone swarm performances in the real world requires planned trajectories to be collision-free and dynamically feasible, respecting actuation and safety constraints. Recent work has increasingly combined learning-based models with model-based control to ensure safety and performance in real-world deployments~\cite{brunke2022safe}. While these methods have shown strong results in single-agent systems, their application to multi-agent swarms is still emerging.

For drone swarms, optimization-based motion planning is a natural choice to enforce constraints explicitly. Coordination strategies are typically classified as centralized, where a global optimization problem resolves conflicts among agents~\cite{augugliaro2012generation, augugliaro_dance_2013, mip_how}, or distributed, where each agent solves a local problem based on partial knowledge~\cite{luis-ral20, soria2021distributed, Adajania2023amswarm}. While both can handle safety and feasibility, distributed approaches scale better with swarm size~\cite{luis2019trajectory}. \rev{Note that a distributed solution is often still run on a central computer, yielding significant scaling benefits from parallelization compared to centralized methods.}
\rev{However, existing distributed methods~\cite{luis-ral20, soria2021distributed, Adajania2023amswarm} are generally limited to point-to-point transitions without fixed timing constraints.}

In this work, we propose a distributed safety filter that extends existing approaches by considering multi-target trajectories with time synchronization, ensuring that drones reach specific positions at fixed times. This formulation enables efficient, real-time enforcement of safety and feasibility during language-driven drone swarm performances.  
\section{The SwarmGPT Framework}
In this section, we present our proposed SwarmGPT framework for generating safe swarm performances from language. An overview of the proposed framework is shown in Fig.~\refwithletter[a]{fig:overview}, which consists of the following core modules: \textit{(i)} a music processor, \textit{(ii)} a language-based choreographer, \textit{(iii)} a motion generator, and \textit{(iv)} an optimization-based safety filter. Details of the individual modules are presented in the respective subsections below. The choreography is executed by a swarm of quadrotors, see Fig.~\refwithletter[b]{fig:overview}.

\subsection{Music Processor}
In the music processor, we analyze the waveform of the raw audio signal to determine a set of beat times to which the performances are synchronized. The beat times are computed based on the spectral-based novelty function of the waveform, following the approach outlined in~\cite{muller2015fundamentals}. Intuitively, the novelty function exemplified in Fig.~\refwithletter[c]{fig:overview} characterizes the distance between successive signal points of the audio, and gives us a measure of how much tones stand out. We define beat times
\begin{equation}
    \Bbb = \{t_0, t_1,...,t_T \}
\end{equation}
as the peaks of the novelty function, where $T\in\mathbb{N}$ corresponds to the total number of beats. To extract the beats, we use standard peak detection tools from \cite{Pauli2020SciPy}. \rev{These extracted timestamps serve as discrete synchronization points for the drone choreography. Specifically, the LLM is permitted to specify motions only at these beat times, ensuring that drone actions are temporally aligned with the musical structure.} The \rev{beat times} are further annotated with their novelty, decibels relative to full scale (i.e., their loudness) and their respective chords. To estimate chords from the audio track, we make use of the hidden Markov model-based approach described in \cite{muller2015fundamentals}. Including this information enables planning performances that are linked more closely to the original audio track. \rev{This annotation enables the planning process to incorporate both rhythmic and expressive elements of the audio, resulting in performances that are closely linked to the audio track.}

\subsection{Language-Based Choreographer}
Given the extracted beat times and audio features, each performance is designed by a \ac{LLM} (specifically GPT-4~\cite{achiam2023gpt} in our case) that defines reference positions for the swarm. The initial prompt introduces the task, provides the music information, and specifies output actions as shown in Fig.~\refwithletter[d]{fig:overview}. Grounding \acp{LLM} in physical embodiments is challenging and can be greatly aided by selecting suitable action modalities \cite{jiang2023vima}. Therefore, we investigate the performance of our framework with two distinct action modalities.

\subsubsection{Waypoints} 
We let the \ac{LLM} generate a series of target coordinates for the beat times for each drone in the swarm. While this allows for a large variety of possible configurations, the \ac{LLM} needs to reason over multiple coordinates to interpret the current swarm configuration. As a consequence, generating coherent geometrical swarm shapes becomes difficult with growing swarm size. 

\subsubsection{Motion Primitives} Alternatively, we let the \ac{LLM} compose performances based on a library of 12 parameterized motion primitives. We augment the initial prompt with a list of available primitives such as \verb|rotate|, \verb|helix|, \verb|spiral| and \verb|wave| with usage examples. A selection of these primitives can be seen in Fig.~\ref{fig:primitives}. 
The \ac{LLM} selects primitives, their parameters, and respective motion start beats $t_{i} \in \Bbb$ and end beats $t_{f} \in \Bbb$. We present a detailed discussion of the motion primitives in the following subsection.

\rev{In case of planning failures such as omitting beat times, missing waypoints for drones, using non-existent primitives or when parameters selected by the \ac{LLM} result in trajectories violating physical limits, the \ac{LLM} receives a failure report and triggers a corresponding self-correction.} In addition to self-corrections, after visualizing the performance in either simulation or experiments, users can modify the swarm choreography directly using natural language via additional reprompting. A full example of the prompts is included in supplementary material S8.

\subsection{Motion Generator}
Generating motions for waypoint actions is straightforward by specifying all desired coordinates and relying on the safety filter to generate physically feasible, collision-free trajectories between points. Our motion generators for primitives for a swarm of $N$ drones are defined as tuples~\cite{Du2018FastAI}:
\begin{equation}
\mathcal{MP}_m = \left( t_{m,i}, t_{m,f}, \{ \bc_{m,n} \}_{n=1}^N,  \mathcal{T}_m(\bc, t) \right),
\end{equation}
where $m\in\Nbb_{[0,M]}$ is the motion primitive index, $t_{m,i} \in \Bbb$ and $t_{m,f} \in \Bbb$ are the start time and the end time of the motion primitive, respectively, $\bc_{m,n}\in\Rbb^3$ is a unique configuration vector associated with a particular drone~$n$ for the motion primitive $m$, and $\mathcal{T}_m(\bc, t) : \Rbb^3 \times [t_{m,i}, t_{m,f}] \to \Rbb^3$ is the position reference trajectory generator. The position reference trajectory for each drone is 
\begin{equation}
    \mathbf{r}_{m,n}(t) = \mathcal{T}_m(\bc_{m,n}, t),\: \forall t\in[t_{m,i}, t_{m,f}].
\end{equation}

\begin{figure}
    \centering
    \subfloat[Wave Primitive]{\includegraphics[width=0.24\textwidth]{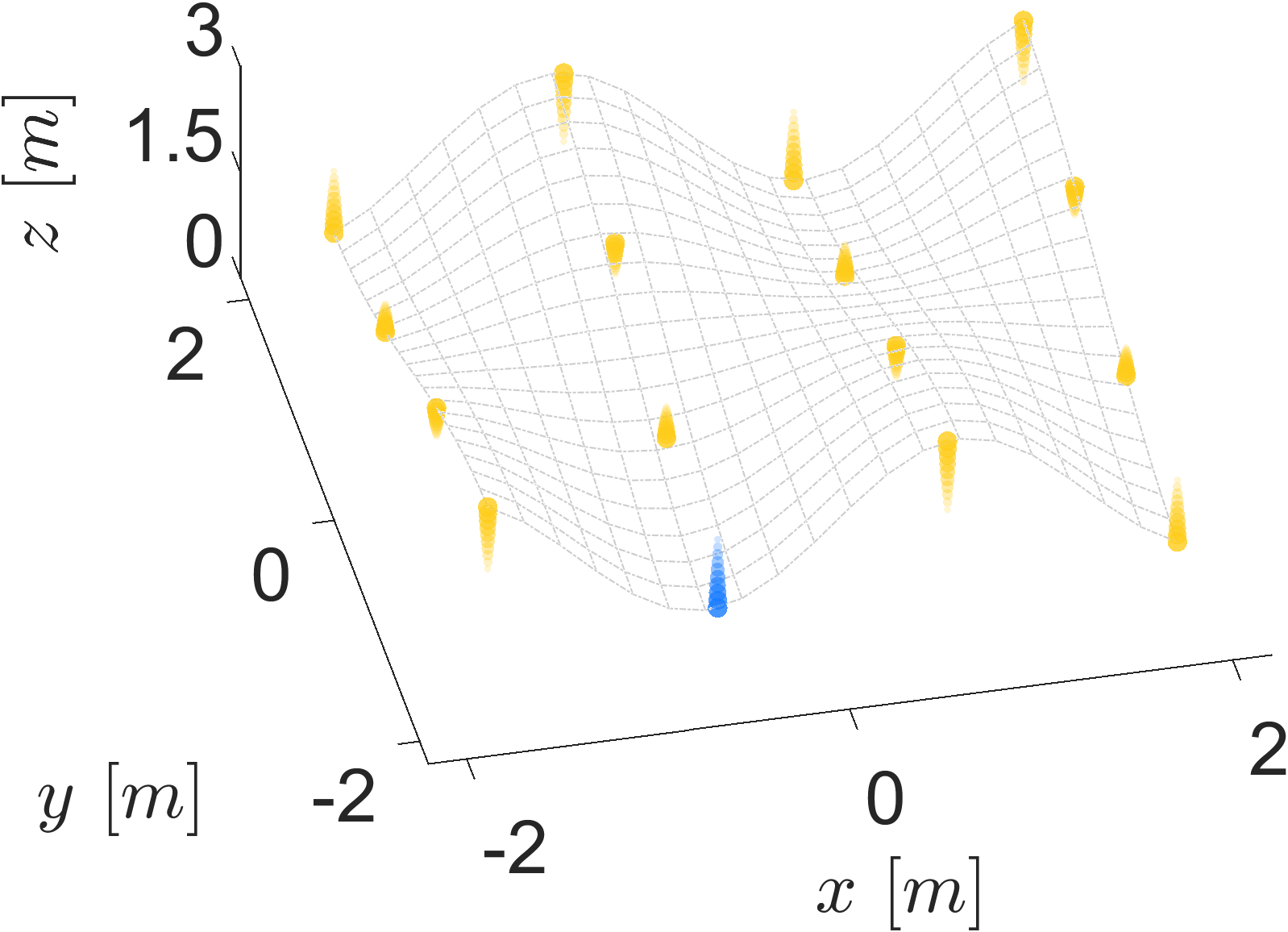} \hspace{-0.1cm}}
    \subfloat[Spiral Primitive]{\includegraphics[width=0.24\textwidth]{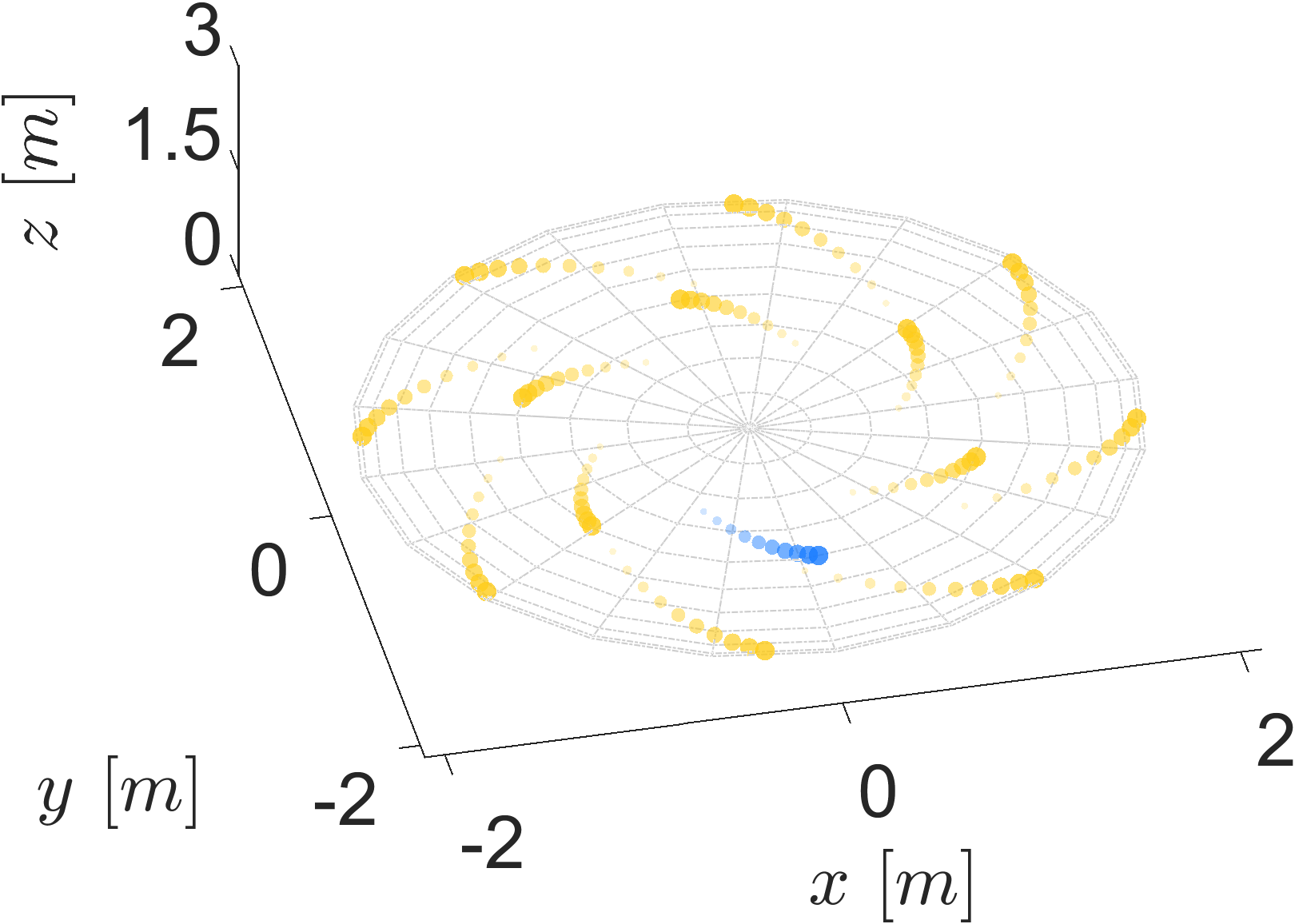}}
    \caption{Visualizations of two example motion primitives in the form of \eqref{eqn:primitive_form}. The largest solid markers show drones’ current positions with trailing tails indicating motion over time; a single blue trajectory highlights one drone’s path.
\textbf{(a)} Wave primitive mimicking a surface wave on a grid-configured swarm, with frequency and amplitude parameters adjustable by the LLM~\cite{Du2018FastAI}.
 \textbf{(b)} Spiral primitive rotating a circular swarm about the $z$-axis while gradually expanding its radius, \rev{configurable by the LLM}.
    }
    \label{fig:primitives}
\end{figure}

There are different ways to define trajectory generators. One general formulation has the following form (with the trajectory index~$m$ dropped for clarity):
\begin{equation}
\label{eqn:primitive_form}
    \mathcal{T}(\bc_n, t) = \mathcal{M}(\bc_n) + \mathcal{T}_\text{per}(\bc_n, t) + \mathcal{T}_\text{poly}(\bc_n, t),
\end{equation}
where $\mathcal{M}(\bc_n)$ is constant with respect to time, and $\mathcal{T}_\text{per}(\bc_n, t)$ and $\mathcal{T}_\text{poly}(\bc_n, t)$ are periodic and polynomial terms defined as follows: 
\begin{align}
\label{eqn:periodic}
    \mathcal{T}_\text{per}(\bc_n, t) &=\sum_{p=1}^P \big( \mathcal{A}_p(\bc_n) \sin(\omega_p t) + \mathcal{B}_p(\bc_n) \cos(\omega_p t) \big),\\
\label{eqn:nonperiodic}
    \mathcal{T}_\text{poly}(\bc_n, t) &=\sum_{q=1}^Q \mathcal{C}_q(\bc_n)\: t^q.
\end{align}
In the general formulation~\eqref{eqn:primitive_form}, the terms $\mathcal{M}(\bc_n)$, $\mathcal{A}_p(\bc_n)$, $\mathcal{B}_p(\bc_n)$, $\mathcal{C}_q(\bc_n)$, $\omega_p$, $P$, and $Q$ are parameters defining the motion primitives. The parameters $\mathcal{A}_p(\bc_n)$, $\mathcal{B}_p(\bc_n)$, $\mathcal{C}_q(\bc_n)$ are drone-dependent amplitude functions for each term. Motion components for some or all drones can be omitted by setting the corresponding amplitudes to zero. Note that, when $\mathcal{C}_q(\bc_n)$ is zero for all $q$ and $n$, we recover the motion primitive generator defined in~\cite{Du2018FastAI}. As shown in~\cite{Du2018FastAI}, with the first two terms alone, diverse periodic behaviours such as rigid body rotation and wave patterns can be achieved for rhythmic drone performances. With the addition of $\mathcal{T}_\text{poly}(\bc_n, t)$, we can further incorporate non-periodic components into the choreography. Two example motion primitives are shown in Fig.~\ref{fig:primitives}. 

Since the equations are difficult to interpret semantically, we generate a library of motion primitives with physically interpretable parameters \rev{that the \ac{LLM} can use} to capture the audio features or incorporate human feedback. For example, we generate the \texttt{rotate} \rev{primitive} by setting the polynomial amplitudes to zero and keeping the height fixed. The \ac{LLM} can specify the desired angular displacement, which we then \rev{convert to} $\omega_p$. Importantly, users do not specify these equations directly but instead issue high-level prompts such as ``move faster,'' which are translated into \rev{primitives} and parameter changes by the \ac{LLM}. \rev{Our motion primitives are similar in spirit to action template design in language-based task planners\cite{singh2023progprompt}.}

\subsection{Optimization-Based Safety Filter}
The language-based choreographer outputs the performance as a set of position reference trajectories $\br_n(t)$ for each drone~$n$ parametrized either by waypoints or through composition of motion primitives. Waypoint references offer no safety guarantees since they do not restrict the LLM's output to collision-free and feasible swarm configurations in any way. Similarly, while individual motion primitives can be designed such that the trajectories $\mathcal{T}_m(\bc_{m,n},t)$ are smooth and collision-free for the intervals $t\in [t_{m,i},t_{m,f}]$, this does not guarantee safe performances. In particular, in between two consecutive motion primitives, the swarm needs to transition dynamically based on their current positions, and these transitions can result in collisions and non-smooth behaviours. 

To certify that trajectories are feasible and safe for deploying to the real-world system, we employ an underlying safety filter that optimizes swarm trajectories for smoothness while adhering to safety constraints. Intuitively, the safety filter allows the drones to follow the designed trajectories where possible and deviate from their trajectory where the proximity to another swarm member necessitates collision avoidance as depicted in Fig.~\refwithletter[e]{fig:overview}. The incorporation of the safety filter allows us to decouple choreography design and the satisfaction of safety constraints for deployment.

We formulate the safety filter as a distributed multi-agent trajectory optimization problem that is solved in a receding horizon manner. As compared to centralized methods~\cite{Augugliaro2013Dance,Du2018FastAI}, distributed methods can be implemented efficiently in real-time and scale up to larger swarms~\cite{luis-ral20,Adajania2023amswarm}. \rev{However, prior works have focused on point-to-point transitions of swarms and do not include timing constraints. Since we require scalable, synchronized motions to multiple swarm configurations in succession, we formulate a distributed safety filter that can handle multiple reference points and respects timing constraints.}

We discretize the reference trajectories at fixed time intervals $\delta t$ based on a pre-specified control frequency and, at each sampling time $t_k=t_0+k \delta t$, each drone $n$ solves an independent optimization problem over a finite prediction horizon $[t_k, t_{k+K}]$:
\begin{equation}
\label{eqn:safety_filter}
    \br^*_{n,k:k+K-1|k} = \pi_\text{safe}\big (\bx_{n,k}, \br_n, \{\bp^*_{j,k:k+K|k-1}\}_{j\in \Jbb} \big ),
\end{equation}
where the notation ``$i|k$'' means the variable at time step~$i$ predicted at time step $k$, ``:'' abbreviates consecutive time steps, $K\in \Nbb$ is the prediction horizon, $\bx_{n,k}=[\bp^T_{n,k},\bv^T_{n,k}]^T$ is a vector containing the current position and velocity of the drone, $\Jbb$ is the set of drone indices with which drone $i$ needs to avoid collision, and $\bp_{j,k:k+K|k-1}^*\in\Rbb^3$ is the predicted position of neighbouring drone $j$ in the swarm based on the optimization from the previous time step. The problem in~\eqref{eqn:safety_filter} is solved for every time step $t_k \in \{t_0, t_0+\delta t, t_0+2 \delta t,..., t_T\}$ for each drone~$n$, and the certified position reference for drone~$n$ at $t_k$ is given by the first element of the optimized variable obtained at each time step: $\br_n^*(t_k)=\br^*_{n,k|k}$.

The safety filter optimization problem $\pi_\text{safe}$ solved for individual drones at time step $k$ is defined as follows (with the drone index $n$ dropped for brevity):
\begin{subequations}
\label{eqn:safety_fillter_optimization}
\begin{align}
\min_{\bu_{\cdot|k}} &\: \sum_{i\in\Kbb}\alpha \left\|\bp_{ i+1|k}^{(D)}\right\|^2 + \beta\left\|\bu_{i|k}^{(D)}\right\|^2 \label{eq:objective}\\
\text{s.t.}&\quad \bx_{k|k} = \bx_k\label{eq:initial_constraint} \\
&\quad  \bu_{k|k}^{(d)} = \bu_{k|k-1}^{(d)}, \:\forall d \in \mathbb{D}\label{eq:continuity}\\
&\quad \bx_{i+1|k} = \mathbf{A}\bx_{i|k}+ \mathbf{B}\bu_{i|k}, \:\forall i \in \Kbb\label{eq:dynamics_constraint}\\
&\quad \bp_{s|k}^{(d)} = \br^{(d)}(t_{s}), \:\forall s \in \mathbb{S}_k,\: \forall d\in \mathbb{D} \label{eq:waypoint_constraint}\\
&\quad\underline{\bp} \preceq \bp_{i|k} \preceq \overline{\bp}, \:\forall i \in \Kbb\label{eq:position_constraint}\\
&\quad\left\Vert \bp_{i|k}^{(1)}\right\Vert^2 \leq \overline{v}^2,  \:\forall i \in \Kbb\label{eq:velocity_constraint} \\
&\quad\underline{f}^2\le \left\Vert \bp_{i|k}^{(2)}+\bg\right\Vert^2  \leq \overline{f}^2, \:\forall i \in \Kbb\label{eq:acceleration_constraint}\\
&\quad\left\Vert \bp_{i|k} - \bp^*_{j,i|k-1}\right\Vert_{\Theta_{j}^{-1}}^2 \geq 1, \:\forall i \in \Kbb, \forall j\in\Jbb, \label{eq:collision_constraint}
\end{align}
\end{subequations}
where $\Kbb = \{k, k+1, ..., k+K-1\}$ is the set of time indices over the horizon, $\alpha,\beta\in \Rbb_{\ge 0}$ are cost function parameters, $\bu_{\cdot|k}$ is the optimization variable parameterized as Bernstein polynomials~\cite{Adajania2023amswarm}, $\bu_{k:k+K-1|k}$ are values of $\bu_{\cdot|k}$ sampled at discrete time steps, the superscript $(\cdot)^{(d)}$ denotes the $d$-th order derivative with respect to time, $\mathbb{D}=\{0,1,...,D\}$ with~$D$ being the desired level of continuity enforced, $\mathbb{S}_k$ is a set of time indices for sampling the reference trajectory, $\bx_{i|k} = [\bp^T_{i|k}, \bv^T_{i|k}]^T$ with $\bv_{i|k} = \bp^{(1)}_{i|k}$, $(\underline{\bp}, \overline{\bp})$ are the position bounds characterizing the size of the flying arena, $\overline{v}$ is the maximum allowable speed, $(\underline{f}, \overline{f})$ are the minimum and maximum mass-normalized collective thrusts generated by the motors, and $\Theta_{j}$ is a diagonal matrix with the diagonal elements characterizing the dimensions of ellipsoidal envelopes for collision and downwash avoidance~\cite{preiss2017downwash}. The matrices, $\mathbf{A}$ and $\mathbf{B}$, are identified system matrices representing the closed-loop dynamics of the quadrotor system~\cite{luis2019trajectory}. Intuitively, the safety filter optimization~\eqref{eqn:safety_fillter_optimization} encourages the certified trajectory $\br^*(t)$ to match the trajectory generated by the language-based choreographer $\br(t)$, while ensuring the trajectory is smooth, feasible, and collision-free for real-world deployment. \rev{Note that \eqref{eq:waypoint_constraint} ensures that drone motions are time synchronized with the trajectory and is enforced for all waypoints within the current horizon.}

The optimization problem in~\eqref{eqn:safety_fillter_optimization} is a non-convex, quadratically constrained quadratic program (QCQP), which can be computationally expensive to solve, especially for large swarms. 
In this work, similar to AMSwarm~\cite{Adajania2023amswarm}, we reparametrize the quadratic constraints in polar form and solve the distributed optimization problem using an alternating minimization scheme. \rev{This approach allows us to efficiently solve the problem in~\eqref{eqn:safety_fillter_optimization} in real-time, without relying on conservative approximations (e.g., linearizing the quadratic constraints as in~\cite{luis-ral20, luis2019trajectory,soria2022distributed}). As compared to ~\cite{Adajania2023amswarm}, our method additionally accounts for the drones' tracking dynamics and enforces continuity in the reference trajectories, which enables more faithful realization of the intended choreography.}

\section{EXPERIMENTAL RESULTS}
Our proposed approach is demonstrated using the Crazyflies~2.1 as shown in Fig.~\ref{fig:intro}. Each Crazyflie is equipped with a controller that converts high-level position references to low-level motor commands. Simulated experiments are conducted with \rev{\texttt{crazyflow}~\cite{crazyflow2025}, a drone swarm simulator modeled after the Crazyflies and successor to \texttt{gym-pybullet-drones}~\cite{panerati2021learning}}. Our hardware setup for real deployment is \rev{further described in Sec. \ref{sec:deployment}.}

\subsection{LLMs as Choreographers}
In our first experiments, we show that SwarmGPT generates drone performances that are synchronized to the given songs and further incorporates human feedback to alter the behaviour of the swarm.

\begin{figure}[t]
    \centering
    \subfloat[\rev{Swarm velocity heatmap over several different songs.}]{\includegraphics[width=\columnwidth, trim={10 20 10 20}, clip]{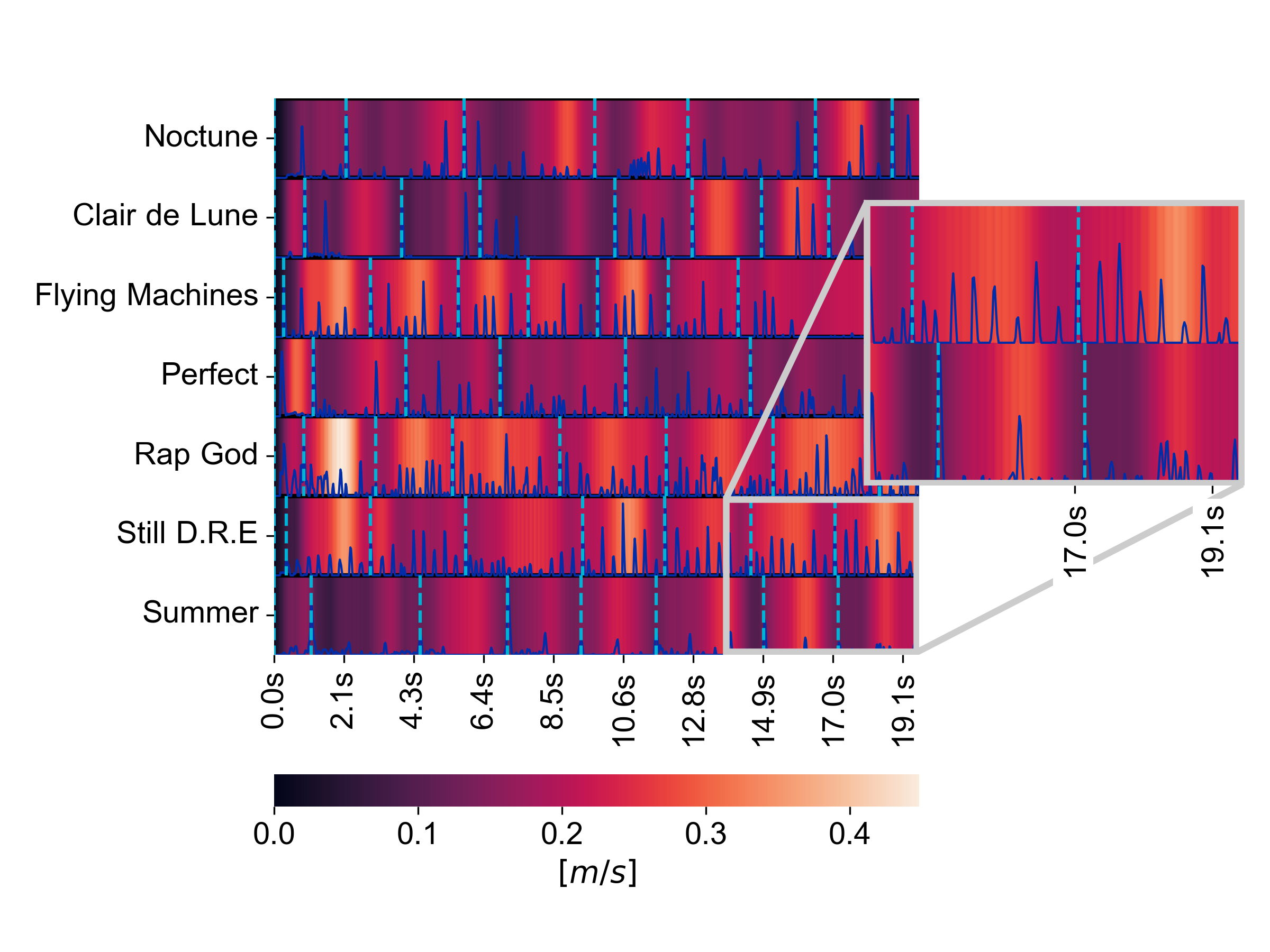}}
    \vfill
    \subfloat[\rev{Speed parameter distribution for a fast (green) and a slow (blue) song.}]{\includegraphics[width=\columnwidth, trim={0 0 0 0}, clip]{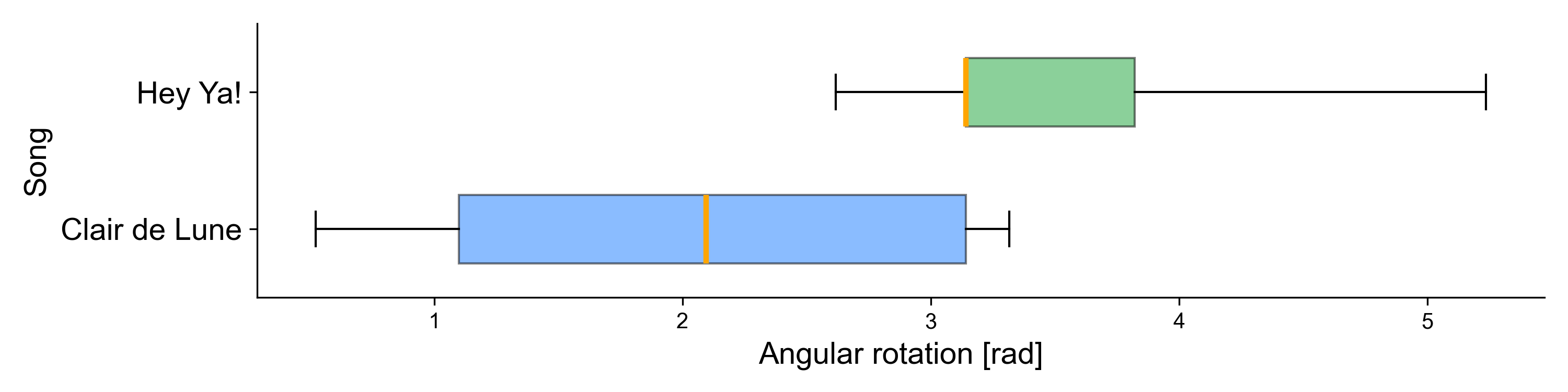}}
    \caption{Matching music to choreography characteristics. \textbf{(a)} Average speed of all drones in the swarm over 10 performances per song. The dark blue line depicts the songs' novelty function, and the dashed light-blue lines denote the beat times. The speed profile is distinct for each song and correlates with the beat times. \rev{\textbf{(b)} Average value of the spiral speed parameter over 25 performances for a slow and a fast song. Boxes denote the 10th and 90th quantiles, with the median in orange. The LLM uses faster speeds for the upbeat song, showing that the music analysis influences the design.}}
    \label{fig:heatmap}
\end{figure}

\subsubsection*{\rev{Performances Match the Music}}
To assess that SwarmGPT produces \rev{distinct and synchronized} performances for different songs, we select seven songs across different genres and create 10 performances each using waypoint actions. Fig.~\ref{fig:heatmap}(a) shows a plot of the average drone speed alongside the beat times and the novelty function derived from music analysis. The plot reveals \rev{that the swarm speed extrema (highlighted in red) are synchronized with the beat times (indicated by dashed light-blue lines), with a distinct pattern for each song. We also observe that performances for classic songs are slower on average than, e.g., the faster rap songs.}

\rev{To further emphasize the connection between music and swarm motion, we design 25 performances using a motion primitive with an interpretable speed parameter for an upbeat song and a slow, classic song each. The distribution of the speed parameter displayed in Fig.~\ref{fig:heatmap}(b) assumes significantly higher values for the lively song, showing that the music analysis is reflected in the design of the LLM.}

\subsubsection*{\rev{User Interaction Through Reprompting}}

\begin{figure}
    \centering
    \includegraphics[width=\linewidth]{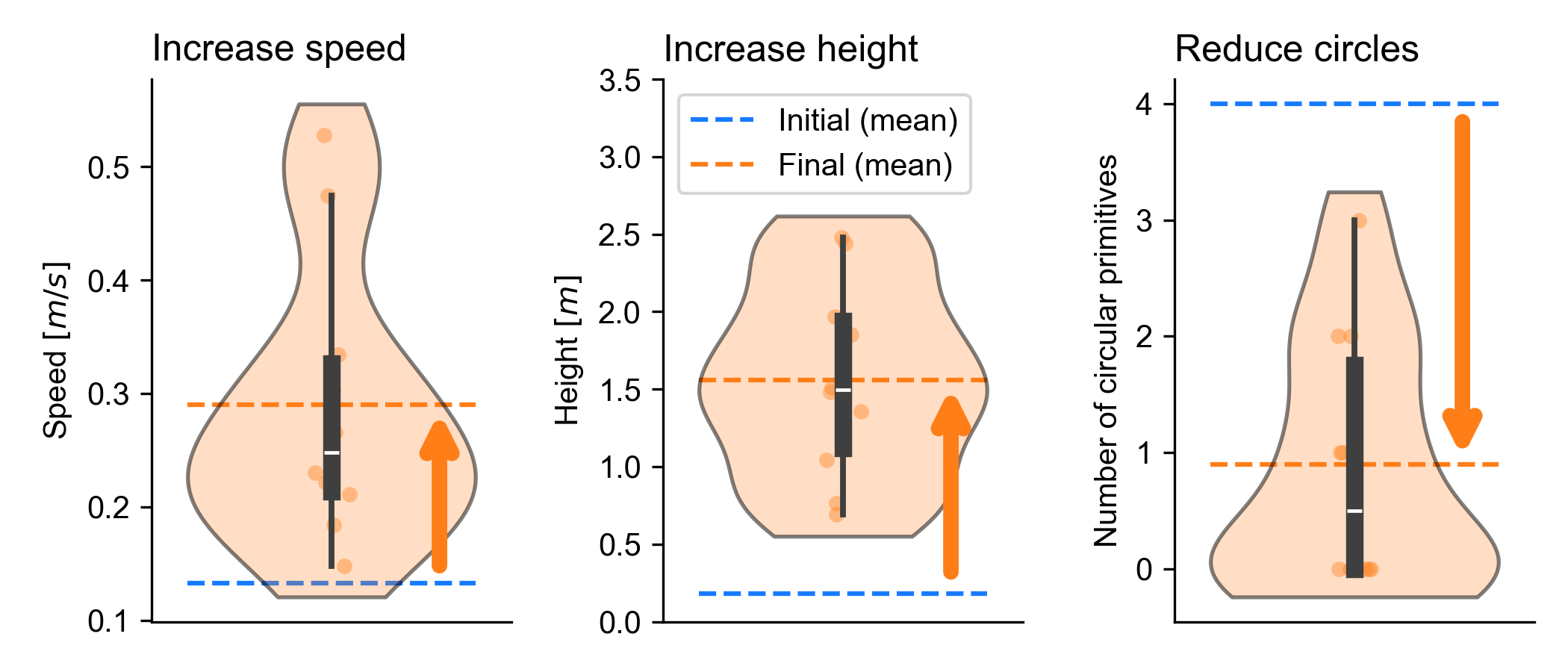}
    \caption{Examples of LLM swarm behaviour modification. \rev{In the user study, 10 participants modified an initial performance (blue) with the goal to increase speed, increase height, or reduce circular motion primitives. The choreography statistics (orange) show participants consistently achieved their goal,} with white being the median, the dashed lines the respective mean values\rev{, and the arrows indicate the desired change}.}
    \label{fig:user_study}
\end{figure}

The LLM can also be reprompted to generate revised versions of the current performance. \rev{We demonstrate this in a user study with 10 participants. Using three pre-designed performances, users are tasked to a) increase the speed, b) increase the height and c) reduce the number of circular motion primitives of the performance. The resulting trajectory statistics in Fig.~\ref{fig:user_study} show that participants consistently achieve their goals. Participants required approximately two minutes to be satisfied with their task. Notably, users were using a wide variety of styles to prompt the choreographer. We thus conclude that SwarmGPT does not only capture the intent of users with its choreographies, but also demonstrates robustness to different prompt formulations.}

\subsection{\rev{Scaling Language-Based Choreographies}}
\begin{figure}
    \centering
    \vspace{10pt}
    \includegraphics[width=\linewidth]{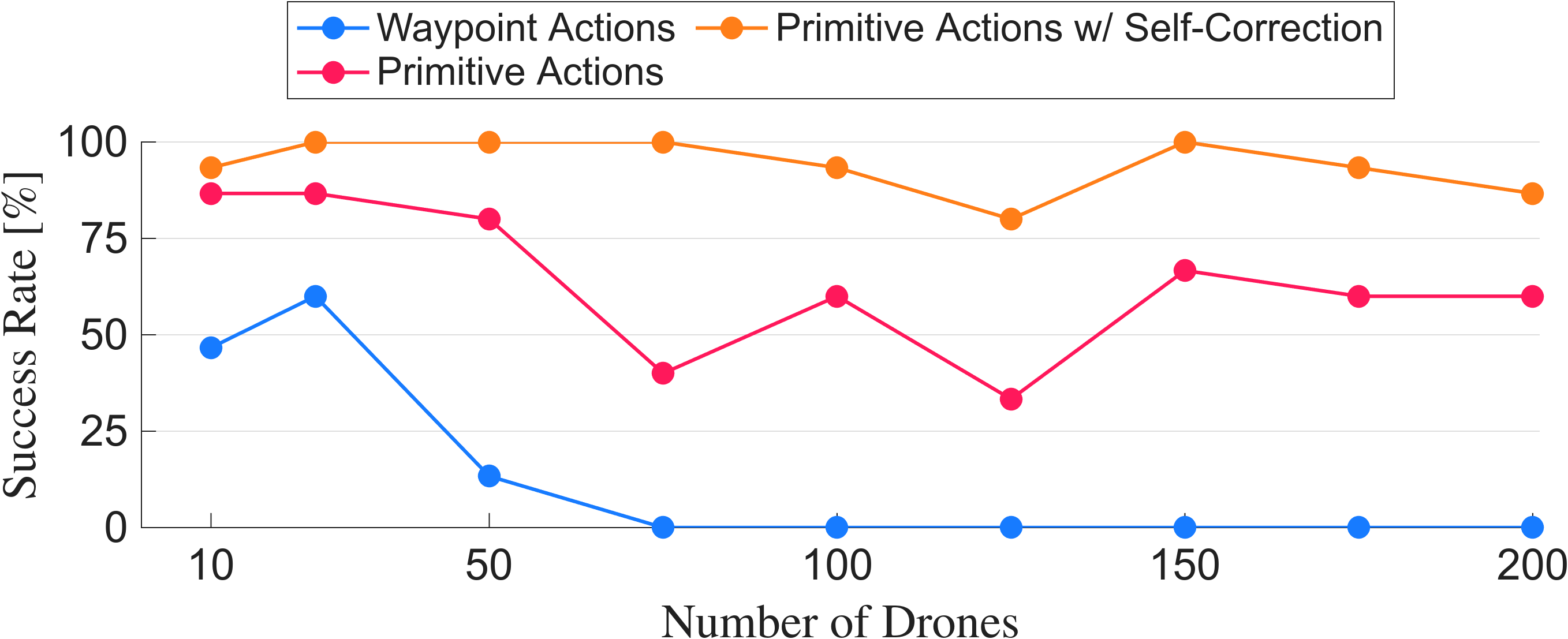}
    \caption{A comparison of success rates for waypoint actions and motion primitives with and without self-correction. We limit self-correction reprompting to two times for each performance. The primitive-based approach leads to significantly higher success rates, with further improvements through self-correction. \rev{Mean response times remain at around 7.5 seconds despite self-corrections, enabling an interactive design process.}}
    \label{fig:success}
\end{figure}

When scaling up the swarm size, the choice of action modalities---waypoints or motion primitives---becomes increasingly significant for successful planning and execution. Specifically, we compare the design success rates of the primitive-based approach with and without self-correction as well as that of the waypoint-based formulation for different swarm sizes. Performances that contain incorrect syntax \rev{after sanitizing responses} or have the same reference simultaneously assigned to two or more drones are considered failures. Each method is evaluated three times on five different songs, which corresponds to a total of 15 performances for each drone swarm size case. As shown in Fig. \ref{fig:success}, the waypoint-based approach yields \rev{approximately} a 50\% success rate for a swarm of \rev{10} drones, but performance degrades to below 20\% for a swarm of \rev{50} drones and eventually drops to 0\%. In contrast, the motion primitive method, which restricts the LLM’s outputs to a set of semantically interpretable parameters, achieves success rates \rev{around} 85\% \rev{that decline initially and then remain around 65\% even for large swarms of 200 drones. Including a self-correction mechanism that automatically reprompts the LLM in case of failures with additional information on the failure type further increases success rates to 85-100\%.} We attribute the rapid deterioration of success rates for waypoint actions to verbose outputs that are prone to syntax and geometric reasoning errors as the number of drones grows, whereas motion primitives remain easier to reason about. \rev{This highlights the importance of using structured responses when scaling towards larger swarms.}

\subsection{Guaranteeing Safety Through Filters} \label{sec:filter}
\begin{figure}
    \centering
    \vspace{10pt}
    \includegraphics[width=\linewidth]{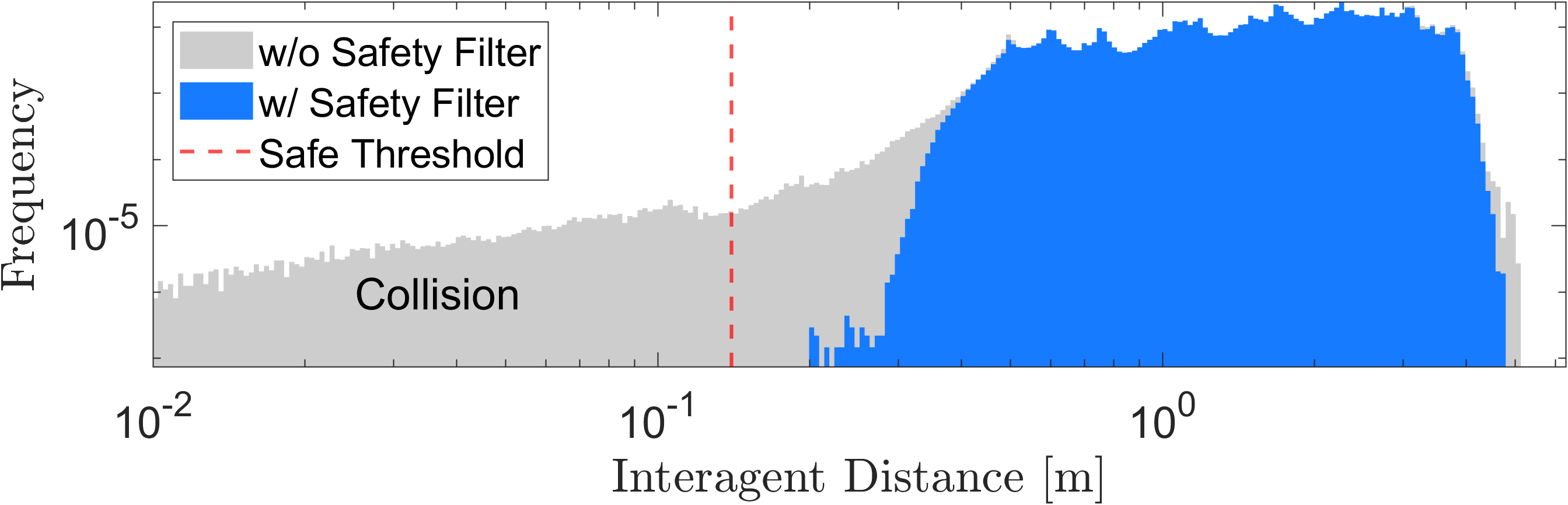}
    \caption{A comparison of interagent distances with and without the safety filter applied. The red dashed line shows the minimum interagent distance to be considered collision-free. The safety filter effectively mitigates collisions that would otherwise occur in language-based choreography designs.}
    \label{fig:safety-filter}
\end{figure}

LLM-generated reference trajectories are inherently unsafe. While motion primitives may be designed for individual feasibility, transitions between them may still be unsafe. Figure~\ref{fig:safety-filter} compares inter-agent distances with and without our safety filter. The results clearly show that the safety filter successfully prevents collisions by resolving unsafe configurations. Evaluations with waypoint actions show similar results.

\begin{figure}[tb]
    \centering
    \vspace{3pt}
\includegraphics[width=\linewidth]{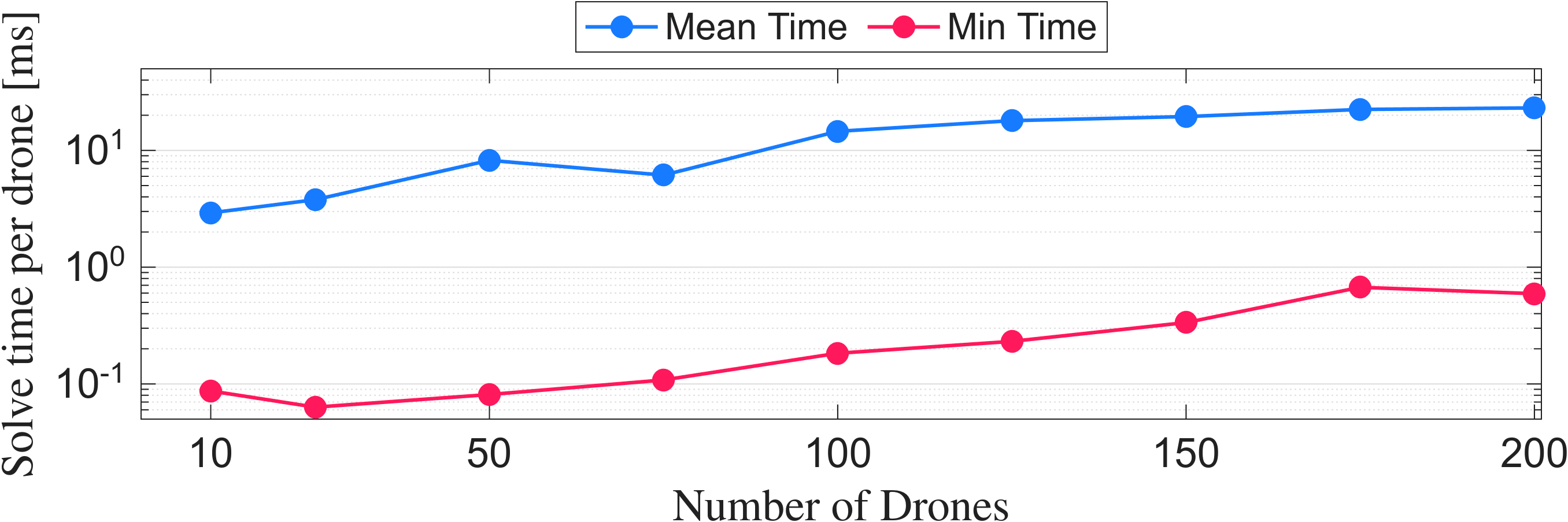}
    \caption{\rev{Per-drone filter solve times for up to 200 drones in milliseconds. As we scale towards larger swarms, the optimization problem becomes more complex and thus per-drone solve times increase despite the distributed formulation. We achieve similar performance to other approaches despite the increased complexity of tracking multiple positions at specific timings.}}
    \label{fig:safety_filter_timings}
\end{figure}

\rev{As the swarm size increases, the safety filter’s optimization becomes more complex. Our distributed formulation solves each drone’s problem independently on a central computer, enabling efficient parallelization and avoiding the high cost of joint optimization. This allows fast solve times, supporting interactive choreography design. Note that ``distributed" refers to the optimization structure—not to inter-drone communication.}
\rev{Computation times of our safety filter are around the same magnitude as other approaches in the literature \cite{luis-ral20, Adajania2023amswarm}. Results of simulating 10 performances each on up to 200 drones are shown in Fig.~\ref{fig:safety_filter_timings}. At 10 drones, the filter on average requires approximately 2.9~ms per drone to solve the problem. This number increases to 23.1~ms for 200 drones. Note that we are solving a more difficult problem compared to prior works, including multiple positions with timing constraints and an additional dynamics model. Our safety filter is implemented in JAX \cite{jax2018github} for ease of integration and scalability and is recomputed at 8~Hz.}

Feasible motion primitives reduce the need for safety filter intervention. Across 15 performances with 20 drones, filtered trajectories deviate on average by only 8.49~cm for primitives, compared to 23.02~cm for waypoint actions, allowing the swarm to more faithfully follow the LLM-designed choreography. \rev{This fidelity is especially important for larger swarms, where infeasible plans may prevent filter convergence.}

\subsection{Real-World Deployment} \label{sec:deployment}
We demonstrate the capabilities of SwarmGPT by deploying performances for swarms of up to 20 drones on the Crazyflies in real-world experiments. 
\rev{Our setup uses a central base station with two antennas and the Crazyswarm platform~\cite{Preiss2017Crazyswarm} to broadcast 10~Hz position commands to drones, which are tracked via Vicon and onboard controllers. Poses are estimated using motion capture and sent from the base station to each drone. Although our filter runs in real time for the deployed swarm sizes, we pre-compute trajectories to eliminate delays and ensure convergence, with optional simulation previews. No inter-drone communication is required, and scaling to larger swarms is possible by adding base stations.}

A long-exposure capture of one performance is included in Fig.~\ref{fig:intro}, and videos of the drone performances can be found at \href{http://tiny.cc/swarmgpt}{\url{http://tiny.cc/swarmgpt}} and in the supplementary materials (S3-S7). In the videos, we give a complete walkthrough of SwarmGPT and showcase designs based on various music genres and the capabilities of the system for modifying swarm behaviours through language instructions (see S2). The performances are accurately tracked by the drones' controllers, with an average tracking error of approximately $4.8 \pm 2.77$~cm. 
\section{CONCLUSIONS}
We introduced SwarmGPT, a framework for generating drone swarm choreographies from natural language. By leveraging the reasoning capabilities of LLMs and an intuitive interface, SwarmGPT enables non-experts to design and refine performances to music. An optimization-based safety filter ensures feasibility by minimally adjusting unsafe trajectories. \rev{We validated our approach in simulation with up to 200 drones and in real-world experiments with up to 20 drones, demonstrating scalability across different swarm sizes and music styles.} 
\rev{While motion primitives help with spatial reasoning and reduce planning failures, they limit flexibility in practice. Future work will explore how to balance structured outputs for scalability with the freedom needed to expand the design space.
Overall, SwarmGPT offers a blueprint for safely integrating LLMs into complex robotic systems and opens the door to broader applications, such as non-expert swarm control in disaster response and search-and-rescue.}

\bibliography{references}
\bibliographystyle{ieeetran}
\appendices
\section{Supplementary Materials}
Supplementary materials include (a) S1, an explainer video of our work, (b) S2, a walk-through video of SwarmGPT, (c) S3-S7, recordings of SwarmGPT performances, and (d) S8 example prompts used. Supplementary materials are accessible at \href{https://tiny.cc/swarmgpt}{\url{https://tiny.cc/swarmgpt}}.




\end{document}